\documentclass[letterpaper]{article} 
\usepackage[preprint]{aaai2027} 
\usepackage[hyphens]{url} 
\usepackage{graphicx} 
\usepackage{natbib} 
\usepackage{caption} 
\usepackage{booktabs}
\usepackage{amsmath}
\title{SANTS: A State-Adaptive Scheduler for
  World Action Models}
\author{
  Yirui Sun\textsuperscript{\rm 1}\equalcontrib,
  Guangyu Zhuge\textsuperscript{\rm 1}\equalcontrib,
  Keliang Liu\textsuperscript{\rm 1},
  Jie Gu\textsuperscript{\rm 1},\\
  Shiqin Dai\textsuperscript{\rm 2},
  Xinyu Bing\textsuperscript{\rm 1},
  Zhongxue Gan\textsuperscript{\rm 1}\corresponding,
  Chunxu Tian\textsuperscript{\rm 1}\corresponding
}
\affiliations{
  \textsuperscript{\rm 1}Fudan University\\
  \textsuperscript{\rm 2}NVIDIA\\
  Emails: \url{sunyr25@m.fudan.edu.cn},
  \url{chxtian@fudan.edu.cn},
  \url{ganzhongxue@fudan.edu.cn}\\
  Project page: \url{https://advanced-robotics-lab.github.io/SANTS/}
}

\makeatletter
\let\aaai@arxiv@original@maketitle\@maketitle
\def\@maketitle{%
  \aaai@arxiv@original@maketitle
  \par\vspace{-0.75em}
  \noindent\begin{minipage}{\textwidth}
    \centering
    \includegraphics[width=\textwidth]{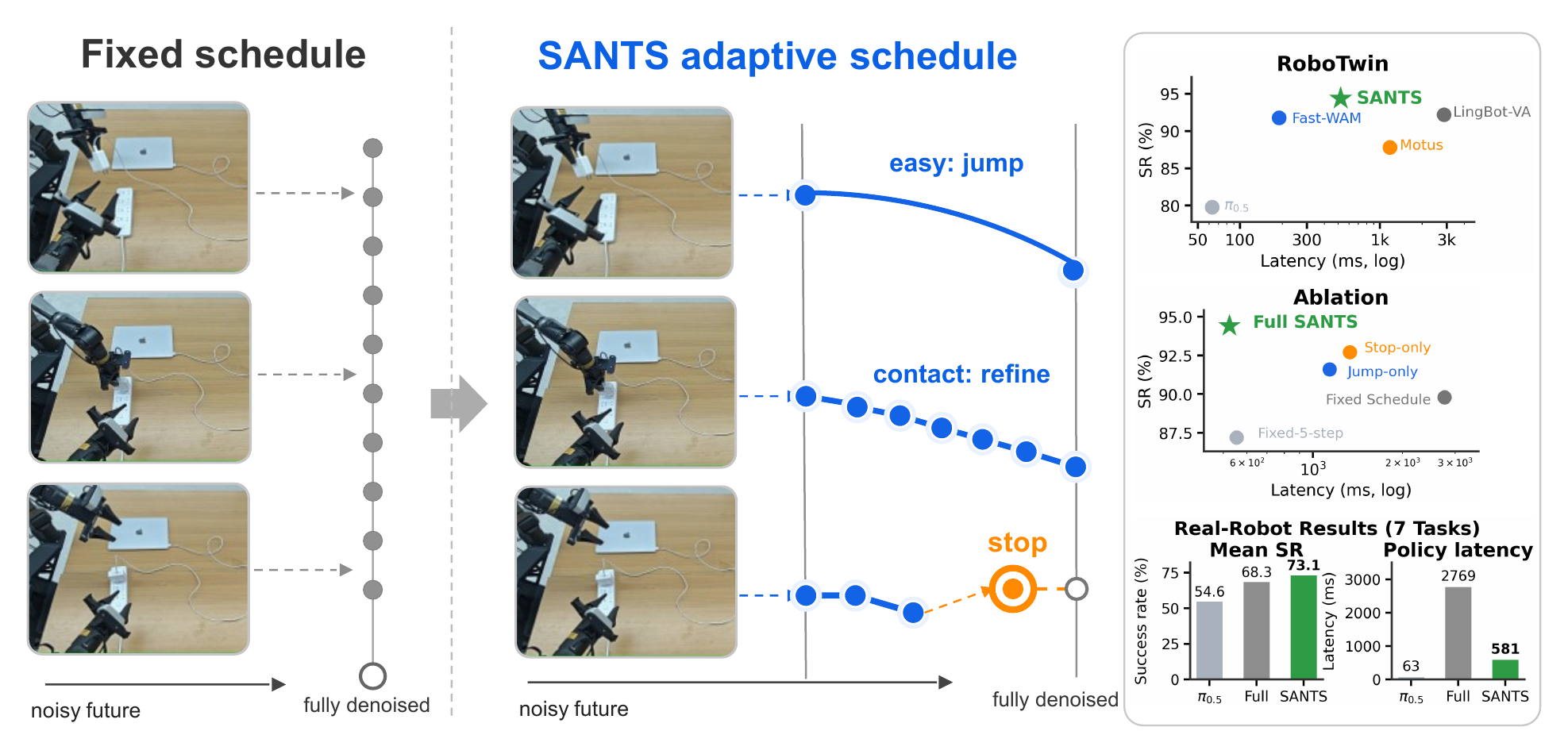}
    \captionof{figure}{Overview of SANTS and main results. The left side contrasts SANTS with a full-denoising WAM: instead of always running to the final denoised state, SANTS selects state-dependent intermediate video states. The right side summarizes success-rate and latency comparisons on RoboTwin 2.0, scheduling ablations, and real-robot experiments.}
    \label{fig:teaser}
  \end{minipage}
  \vspace{0.75em}
}
\makeatother

\begin{document}

\maketitle

\begin{abstract}
World Action Models (WAMs) improve robot manipulation by using video-based future representations to condition action generation. In pixel-space WAMs, however, the best action condition is not necessarily the fully denoised video. Controlled denoising-depth scans show that video refinement can reduce action error up to a state-dependent point, after which the gain may saturate or even reverse when late predictions become less action-relevant or physically unreliable. This suggests that action generation should use a state-dependent point along the video noise trajectory rather than a fixed terminal denoising depth. We introduce State-Adaptive Noise Trajectory Scheduler (SANTS), a lightweight scheduler for video-to-action diffusion policies. At each video decision point, SANTS reads the current video-state representation and noise level, then jointly predicts a cumulative stopping hazard and a relative noise-progression ratio. SANTS is post-trained with a path-level reward computed after the frozen action branch generates the final action chunk, so the scheduler is optimized for downstream action quality rather than intermediate video fidelity, while redundant video-state updates are explicitly penalized. Experiments show that SANTS reaches \(94.4\%\) overall success on RoboTwin 2.0 and \(73.1\%\) average success across seven real-robot tasks, while reducing latency by \(81.7\%\) and \(79.0\%\) relative to full video denoising, respectively. These results indicate that adaptive selection along the video noise trajectory can preserve the control benefits of WAM-style future reasoning while removing much of its redundant inference cost.
\end{abstract}


\section{Introduction}

World Action Models (WAMs) have recently emerged as a promising paradigm for robot control by coupling future visual prediction with action generation~\citep{wang2026worldactionmodelsfrontier,zhu2025unifiedworldmodelscoupling,li2026causalworldmodelingrobot,ye2026world}. Instead of mapping observations directly to actions, WAMs expose an intermediate future-evolution interface, allowing large video models to transfer spatiotemporal priors about object motion, contact changes, and scene dynamics to manipulation policies~\citep{kim2026cosmospolicyfinetuningvideo,hu2025videopredictionpolicygeneralist,pai2025mimicvideovideoactionmodelsgeneralizable,ma2026dit4ditjointlymodelingvideo,bi2025motusunifiedlatentaction}. This design has shown strong potential for generalizable control, especially in long-horizon and visually diverse manipulation tasks. However, its test-time execution remains bottlenecked by the video denoising process: generating a future visual condition typically requires many iterative video updates, making inference slower than direct action policies~\citep{yuan2026fastwamworldactionmodels}. More importantly, the value of this future video condition is not determined by visual fidelity alone. For fine-grained manipulation, such as grasping, alignment, insertion, and placement, the action branch requires a representation that preserves contact-relevant and motion-relevant cues, while unnecessary or inaccurate late-stage video refinement can increase latency and may even mislead action generation~\citep{yuan2026fastwamworldactionmodels,zhang2026learningvisualfeaturebasedworld}.

This observation motivates a more precise question: for a pixel-space WAM, which intermediate video representation should be used to condition action generation? Recent video-action policies already suggest that intermediate video features and predictive visual representations can provide useful action conditions~\citep{hu2025videopredictionpolicygeneralist,pai2025mimicvideovideoactionmodelsgeneralizable,ma2026dit4ditjointlymodelingvideo,ye2026world}. Through controlled scans over video denoising depth, we find that deeper denoising generally improves action prediction when the predicted future remains physically and semantically reliable, suggesting that video denoising does provide useful action-conditioning signals. Yet this improvement exhibits clear diminishing returns and state dependence. Coarse states often obtain sufficient action cues after shallow denoising, whereas contact-rich or alignment-sensitive states benefit from deeper future evolution. We also find exceptions: when late video predictions deviate from the underlying physical dynamics, further denoising can make the resulting condition less useful for action generation. These findings suggest that the desired condition for action generation is neither the noisiest intermediate state nor the fully denoised video, but a state-dependent point along the video noise trajectory.



A fixed video denoising schedule is not designed to select such a state-dependent point. It predetermines both the terminal denoising depth and the noise-level grid before observing whether the current state already contains sufficient action cues~\citep{sabour2024alignstepsoptimizingsampling,esteves2026spectrallyguideddiffusionnoiseschedules}. This couples two decisions that should depend on the state: when to stop refining the future-video representation, and how large an integration step to take along the noise trajectory if further denoising is still useful. We propose State-Adaptive Noise Trajectory Scheduler (SANTS), a lightweight scheduler for pixel-space WAMs that turns fixed video denoising into state-adaptive noise-trajectory control. At each video decision point, SANTS reads the current video-state representation and noise level, and predicts two complementary quantities. A cumulative hazard-based stopping head accumulates termination evidence along the trajectory and decides whether the current intermediate video representation is already sufficient for action generation. A relative progression head then predicts how far to progress along the noise axis when denoising continues. This allows SANTS to skip low-value regions when the action intent is clear while preserving finer updates for contact-rich or alignment-sensitive states.

Learning this scheduler is nontrivial because the best intermediate video representation has no explicit label and cannot be supervised by video fidelity alone. We therefore train SANTS as a trajectory policy with a path-level reinforcement signal, following the broader idea of using downstream rewards to trade off denoising quality and inference cost~\citep{yu2025d3pdynamicdenoisingdiffusion}. A sampled noise trajectory is evaluated only after the frozen action branch generates the final action chunk, so the reward directly reflects downstream action quality while penalizing unnecessary video updates. This objective encourages the scheduler to retain video updates only when they provide action-relevant gains, rather than when they merely improve visual reconstruction. During training, we update only the lightweight scheduler and keep the WAM backbone, video head, action head, and action denoising process frozen. As a result, SANTS requires no manually annotated stopping labels, introduces minimal additional parameters, and can be attached to compatible WAMs as a plug-in module to improve both fine-grained action accuracy and inference efficiency.

Our contributions are:
\begin{itemize}
  \item We identify state-dependent intermediate video representation selection as a key inference problem in pixel-space WAMs, supported by controlled denoising-depth scans.

  \item We propose SANTS, a lightweight scheduler that jointly controls stopping depth and relative noise progression for video denoising.

  \item On RoboTwin 2.0~\citep{chen2025robotwin20scalabledata} and real robots, SANTS yields a better success--latency tradeoff, with trajectory analyses and ablations supporting the adaptive scheduling mechanism.
\end{itemize}

\section{Related Work}

\textbf{Evolution of World Action Models.} Recent studies suggest that future-oriented prediction in WAMs can improve zero-shot action generation and cross-task transfer, as it enables policies to better capture the dynamic constraints underlying manipulation tasks~\citep{wu2023unleashinglargescalevideogenerative,cheang2024gr2generativevideolanguageactionmodel,cen2025worldvlaautoregressiveactionworld,ye2026world,feng2026harmowamharmonizinggeneralizableprecise}. Recent WAMs follow several directions. One line of work adopts action-conditioned world models, where candidate actions are rolled out into predicted future outcomes and then selected through planning, search, or value estimation~\citep{finn2017deep,kim2026cosmospolicyfinetuningvideo,li2025worldevalworldmodelrealworld,yang2026riseselfimprovingrobotpolicy}. Another line treats future prediction as an intermediate goal: the model first predicts future images, object motion, or state changes, and then recovers the robot action required to realize the predicted future through inverse dynamics or action inversion modules~\citep{wen2024vidmanexploitingimplicitdynamics,routray2026vipravideopredictionrobot,li2026imaginedfuturesexecutableactions,feng2026harmowamharmonizinggeneralizableprecise}. A third line focuses on the intermediate representations formed during prediction, which serve as action-ready features for the action branch~\citep{hu2025videopredictionpolicygeneralist,li2026causalworldmodelingrobot,liu2026oawamobjectaddressableworldaction,bi2025motusunifiedlatentaction,motubrainteam2026motubrainadvancedworldaction,su2026worldguidanceworldmodeling,luo2026beingh07latentworldactionmodel}.

\textbf{Efficient and Action-Useful WAM Inference.} Many WAMs rely on diffusion-based video prediction, which introduces an inherent trade-off between inference efficiency and prediction fidelity. General-purpose diffusion models are commonly accelerated by few-step samplers, high-order ODE solvers, or learned noise schedules, indicating that denoising does not have to follow a fixed uniform trajectory~\citep{song2020denoising,lu2022dpm,lu2025dpm,zheng2023dpmsolverv3improveddiffusionode,sabour2024alignstepsoptimizingsampling,esteves2026spectrallyguideddiffusionnoiseschedules}. However, these methods are primarily optimized for image quality, reconstruction error, or generative consistency, and do not directly address the utility of denoising for action generation. Recent WAM studies have begun to revisit the video denoising process from the perspective of action utility, suggesting that the action branch does not always need to wait for a fully denoised future video. Instead, intermediate denoising features, compact conditioning representations, or future-aware latent interfaces can already provide action-relevant information about object motion, contact changes, and scene evolution, and have been exploited to reduce inference cost~\citep{hu2025videopredictionpolicygeneralist,su2026worldguidanceworldmodeling,luo2026beingh07latentworldactionmodel,yuan2026fastwamworldactionmodels,guo2026unified4dworldaction,jia2026video2actdualsystemvideodiffusion,ye2026gigaworldpolicyefficientactioncenteredworldaction,motubrainteam2026motubrainadvancedworldaction}. Nevertheless, existing methods typically assume a fixed denoising depth or a fixed feature extraction point. How to select action-useful intermediate representations without fully depending on final video fidelity remains an important question for WAM inference.

\section{Method}

\begin{figure*}[t]
\centering
    \includegraphics[width=0.78\textwidth]{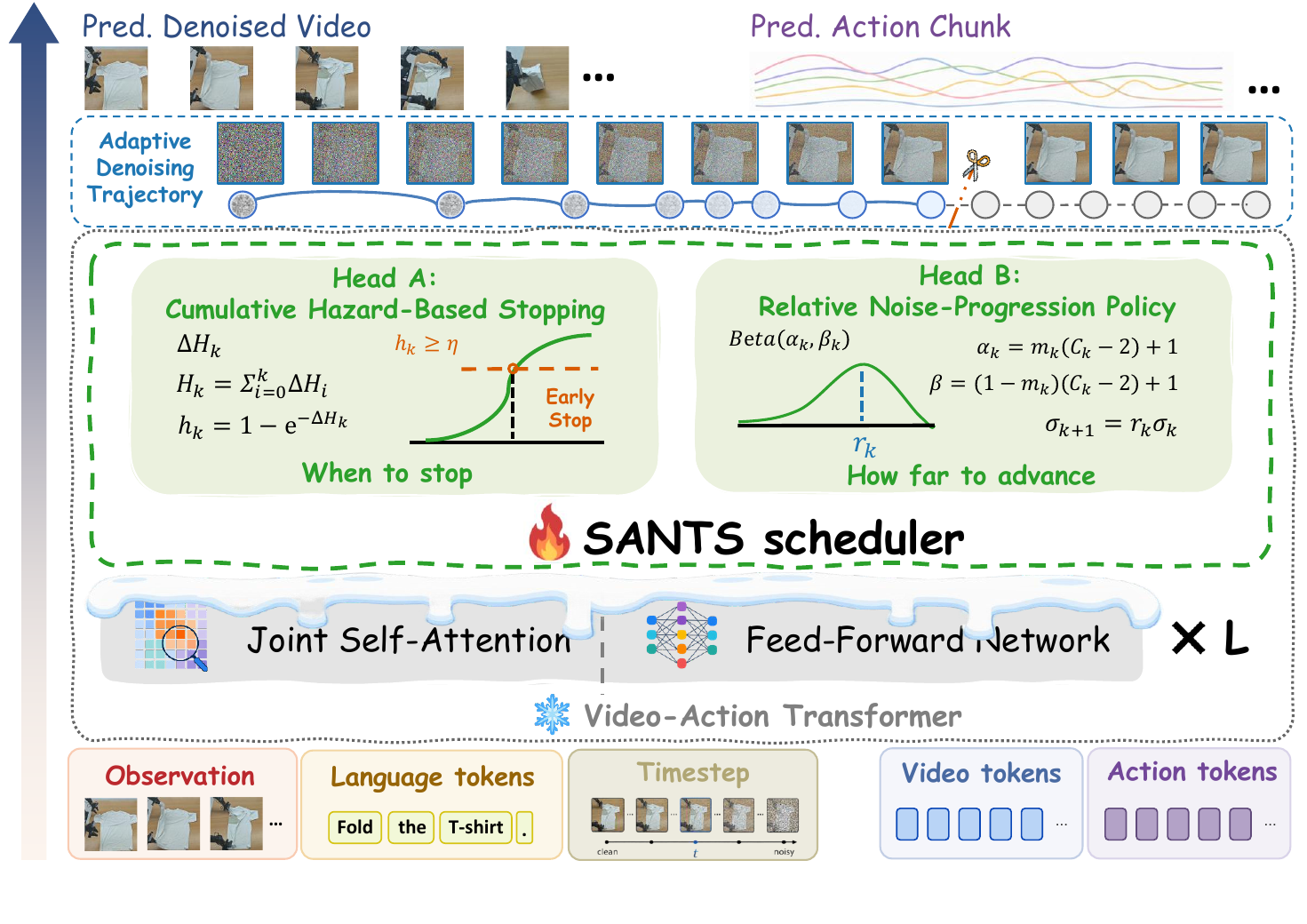}
\caption{Overview of SANTS. SANTS is attached to a frozen video--action diffusion policy. During video denoising, it reads the current video-state representation and noise level, and jointly predicts a stopping hazard and a relative noise-progression ratio to select a state-dependent intermediate video representation for action generation.}
\label{fig:sants-overview}
\end{figure*}

\subsection{Frozen Video--Action Policy Interface}
\label{sec:method-policy-interface}

As shown in Figure~\ref{fig:sants-overview}, SANTS builds on a pretrained video-to-action diffusion policy rather than introducing a new robot-policy backbone. Let \(f_{\phi}\) denote the frozen video--action Transformer. At noise level \(\sigma_k\), the policy maintains a video latent \(x_k\), and each video-denoising forward pass predicts the flow field used to update this latent. We also extract the pooled hidden representation of the video tokens from the same forward pass and denote it by \(z_k\), which serves as the scheduler state. When the scheduler stops at terminal video state \(x_{\tau}\) and terminal noise level \(\sigma_{\mathrm{term}}\), the action branch generates an action chunk \(\hat{a}_{\tau}\) conditioned on the selected terminal video representation.

This interface decouples scheduler learning from policy learning. SANTS observes only \((z_k,\sigma_k)\) and does not update the Transformer blocks, video head, action head, or action denoising process. Thus, different scheduling policies are compared under the same policy capacity, and the learned scheduler can be attached to any compatible video--action diffusion policy that exposes intermediate video features and supports nonuniform noise-level updates.

Based on this interface, SANTS controls the video noise trajectory through two outputs at each decision point: a hazard increment \(\Delta H_k\) for stopping and a relative noise-ratio distribution for continuation. If the policy stops, video integration terminates and the selected terminal representation is passed to the action branch; otherwise, the sampled progression ratio advances the noise level to \(\sigma_{k+1}\). We next define the stopping mechanism, relative noise-progression policy, path-level reward, and scheduler training.

\subsection{Stopping Hazard and Relative Noise Progression}

\paragraph{Cumulative hazard-based stopping.}
We formulate early termination as a sequential decision process along the sampled video denoising path. Let \(k\) denote the current scheduler decision index, and let \(T\) be the first decision point at which video integration terminates. Reaching decision index \(k\) is equivalent to \(T\ge k\); conditioned on this event, \(T=k\) means stopping at the current decision point, and \(T>k\) means continuing. The stopping head first maps the current scheduler state \((z_k,\sigma_k)\) to a scalar score and converts it into a nonnegative hazard increment,
\begin{equation}
\Delta H_k = \mathrm{softplus}(g(z_k,\sigma_k)),
\label{eq:sants-hazard-inc}
\end{equation}
where \(\Delta H_k\ge 0\). The increments are accumulated along the path as
\begin{equation}
H_k = \sum_{i=1}^{k}\Delta H_i,
\label{eq:sants-hazard-cum}
\end{equation}
which represents the termination evidence accumulated up to decision point \(k\). We then define the cumulative stopping probability
\begin{equation}
F_k = P(T\le k) = 1-e^{-H_k}.
\label{eq:sants-stop-cdf}
\end{equation}
The corresponding conditional stopping probability is
\begin{equation}
h_k = P(T=k \mid T\ge k) = 1-e^{-\Delta H_k}.
\label{eq:sants-stop-prob}
\end{equation}
During training, \(h_k\) defines the stochastic stop/continue policy and is used to compute the trajectory log-probability. During deployment, SANTS instead uses the monotonic cumulative probability \(F_k\) as the deterministic termination criterion, stopping once \(F_k\) exceeds a threshold \(\eta\). The cumulative-hazard parameterization keeps the local training probability \(h_k\) and the global deployment evidence \(F_k\) consistent under the same path decomposition.

\paragraph{Relative noise-progression policy.}
Stopping determines when to end video integration, but a continuing path must also decide how far to move along the noise axis. If continuation can only move to the next fixed noise level, all states share the same discrete integration grid. This prevents the scheduler from skipping low-value updates when the action intent is already clear, and from preserving finer video updates when contact or future evolution is uncertain. SANTS therefore models continuation as a relative noise ratio,
\begin{equation}
r_k \sim \mathrm{Beta}(\alpha_k,\beta_k), \qquad \sigma_{k+1}=r_k\sigma_k,
\label{eq:sants-jump-ratio}
\end{equation}
where \(r_k\in(0,1)\) is the relative noise retained for the next decision point, and \(1-r_k\) is the relative progression distance. A value of \(r_k\) close to \(1\) gives a conservative noise progression, while a smaller \(r_k\) gives a more aggressive progression. This formulation lets the scheduler choose the next integration time in continuous noise space instead of being restricted to a preset grid.

To share one parameterization between stochastic exploration in training and deterministic progression at deployment, the scheduler predicts the mode \(m_k\in(0,1)\) and concentration \(c_k>2\), rather than unconstrained Beta parameters. We set
\begin{equation}
\alpha_k = m_k(c_k-2)+1,\qquad \beta_k=(1-m_k)(c_k-2)+1.
\label{eq:sants-jump-param}
\end{equation}
This guarantees \(\alpha_k>1\) and \(\beta_k>1\), yielding a unimodal Beta distribution whose mode is \(m_k\). During training, we sample \(r_k\) from Eq.~(\ref{eq:sants-jump-ratio}) to preserve path exploration; at deployment, we use \(m_k\) directly as the deterministic progression ratio. For an arbitrary selected noise pair \((\sigma_k,\sigma_{k+1})\), we use the same first-order flow-matching update as the base sampler~\citep{liu2022flow},
\begin{equation}
x_{k+1}=x_k+\hat v_{\phi}(x_k,\sigma_k,c)(\sigma_{k+1}-\sigma_k),
\label{eq:sants-continuous-update}
\end{equation}
where \(c\) denotes the observation, language, and history conditions. Each scheduler decision requires one video forward pass to obtain the current state. If the scheduler stops at decision point \(k\), this decision consumes a video forward pass but does not produce a new video-state update; the update count increases only when the scheduler continues and applies the integration update.

\subsection{Path-Level Reward and Scheduler Post-Training}
\label{sec:method-training}

The optimal intermediate video representation has no explicit label and cannot be supervised by video fidelity alone. We train SANTS as a trajectory policy with a path-level reward evaluated after the frozen action branch generates the final action chunk. This makes all stopping and progression decisions optimize downstream action quality rather than intermediate video reconstruction.

For a sampled trajectory \(\tau\), the reward compares the generated action chunk \(\hat{a}_{\tau}\) with the demonstration action \(a^{\star}\) using event-aware sequence and temporal-difference action errors, \(L_{\mathrm{seq}}^{\tau}\) and \(L_{\Delta}^{\tau}\). These errors are converted into normalized action-quality gains and penalized by the number of executed video-state updates:
\begin{equation}
R(\tau)=
\sum_{m\in\{\mathrm{seq},\Delta\}} w_m\,
\Phi_m\!\left(L_m^{\tau},L_m^{\mathrm{lo}},L_m^{\mathrm{hi}}\right)
-\lambda_c\,\Psi(n_{\tau}),
\label{eq:sants-reward-compact}
\end{equation}
where \(n_{\tau}\) is the number of actual video-state updates. The reference errors \(L_m^{\mathrm{lo}}\) and \(L_m^{\mathrm{hi}}\) are computed from a shallow-denoising path and a full-denoising path for the same input, allowing the reward to measure recovered action quality relative to sample difficulty. If the full-denoising anchor provides little or no improvement over the shallow anchor, the reward suppresses positive gain, enabling the scheduler to learn early termination on plateaued or non-monotonic cases. Thus, the reward is not a simple action-error objective: it compares paths under the same input, emphasizes action-critical frames, and trades recovered control quality against the marginal cost of traversing the denoising trajectory. Additional gating, clipping, key-frame weighting, and update-dependent cost details are provided in Supplementary Section C.

During post-training, we update only the scheduler and keep the video--action backbone, video head, action head, and action denoising process frozen. The resulting trajectory policy is optimized with PPO~\citep{schulman2017proximal} using Eq.~(\ref{eq:sants-reward-compact}), with a small reference regularization term to stabilize early training. Scheduler implementation details and target-domain tuning settings are provided in Supplementary Section B.

\section{Experiments}

The experiments answer three questions. We first use an offline diagnostic to test whether action utility along the video denoising trajectory is state dependent, then evaluate closed-loop success and latency in simulation and on real robots, and finally ablate the stopping and progression decisions in SANTS.

\subsection{Offline Diagnostic and Scheduling Behavior}
\label{sec:exp-denoising-depth}
\label{sec:exp-budget-trajectory}

Before closed-loop evaluation, we use an offline depth diagnostic to test the central premise of SANTS: whether a fully denoised future video is always the most useful action condition. Specifically, the diagnostic asks whether action utility peaks at different denoising depths for different states; if so, video generation should be treated as a state-dependent scheduling process rather than a fixed trajectory towards full denoising. We fine-tune the video--action backbone on RoboTwin 2.0 target-domain data~\citep{chen2025robotwin20scalabledata} while sampling intermediate noise levels from the shifted flow-matching schedule~\citep{liu2022flow}, so that the action branch observes video states throughout the denoising trajectory rather than only at a fixed endpoint. We then freeze all policy parameters and scan seven denoising depths \(d\in\{4\%,16\%,32\%,48\%,64\%,80\%,100\%\}\) on \(500\) randomly sampled, manually phase-annotated diagnostic segments. The annotations use only the original demonstration videos and robot motion, without access to depth-scan losses. We label approach, free-space motion, and transport as coarse phases, and contact, alignment, grasping, insertion, and placement as fine phases. For each segment, we fix the initial video and action noise and report masked action MSE relative to its \(4\%\)-depth baseline. This paired construction changes only the video denoising depth within each segment, while the normalisation makes results comparable across segments with different absolute action scales.

\begin{figure}[!t]
\centering
\includegraphics[width=0.88\linewidth,trim=0 0 392bp 0,clip]{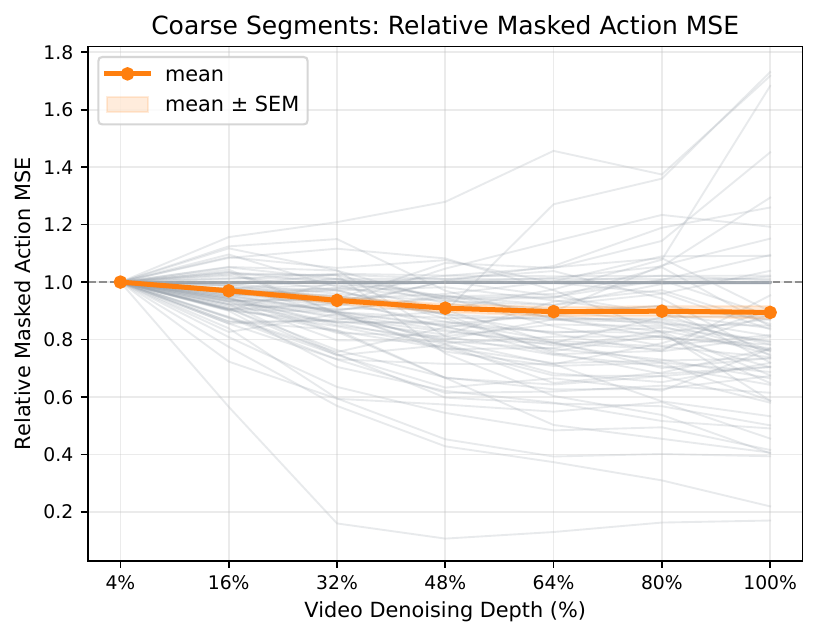}\par
\smallskip
\includegraphics[width=0.88\linewidth,trim=392bp 0 0 0,clip]{figures/sants_denoising_depth_diagnostic.pdf}
\caption{Relative masked action MSE across video denoising depths. The upper and lower panels show coarse and fine segments. Coloured curves and shaded bands report the group mean and its standard error; each grey curve tracks one individual segment. Errors are normalised to each segment's \(4\%\)-depth value, and the dashed line at \(1.0\) marks no change.}
\label{fig:denoising-depth-diagnostic}
\end{figure}

Figure~\ref{fig:denoising-depth-diagnostic} shows rapid error reduction at low-to-moderate depths followed by diminishing returns. Coarse-phase error decreases from \(1.00\) to \(0.89\), whereas fine-phase error reaches \(0.67\), indicating that contact-sensitive motion benefits more, on average, from refined video conditions. However, the individual segment trajectories show frequent plateaus and error increases in both groups. Thus, the coarse--fine grouping explains an average difference in the benefit of additional denoising, but it does not identify the best terminal depth for a particular state. For segments whose action error increases at later depths, visual inspection shows that the predicted future can become physically or dynamically inconsistent, for example by placing the gripper and object in implausibly overlapping configurations. In such cases, additional denoising produces a more refined video prediction but a less useful condition for action generation. These observations provide the empirical motivation for SANTS: inference should select an action-useful intermediate video condition for each state rather than defaulting to a fully denoised future.

\begin{table}[t]
\centering
\small
\setlength{\tabcolsep}{5pt}
\begin{tabular}{lcc}
\toprule
Metric & Coarse & Fine \\
\midrule
100\% not best & 55.65\% & 41.33\% \\
Adjacent increase & 87.90\% & 77.49\% \\
Oracle mean & 0.8433 & 0.6084 \\
Fixed 100\% mean & 0.8943 & 0.6571 \\
\bottomrule
\end{tabular}
\caption{Offline depth-scan statistics. ``100\% not best'' denotes segments whose minimum error occurs before full denoising, and ``Adjacent increase'' denotes segments with an error increase between consecutive scanned depths. ``Oracle mean'' selects the lowest scanned error independently for each segment, whereas ``Fixed 100\% mean'' always uses the fully denoised endpoint.}
\label{tab:app-depth-diagnostic-stats}
\end{table}

Table~\ref{tab:app-depth-diagnostic-stats} quantifies this sample-level non-monotonicity. Full denoising is not the best scanned depth for \(55.65\%\) of coarse segments and \(41.33\%\) of fine segments; \(87.90\%\) and \(77.49\%\), respectively, contain at least one adjacent error increase. The adjacent-increase rate captures local non-monotonicity even when the fully denoised endpoint achieves the lowest overall error. Its high frequency shows that action-prediction error does not decrease monotonically with denoising depth. Selecting the best scanned depth independently for each segment reduces mean error from \(0.8943\) to \(0.8433\) for coarse phases and from \(0.6571\) to \(0.6084\) for fine phases.

This oracle comparison ignores computation and is not the objective of the learned scheduler; it only measures the action-error reduction available from selecting the best scanned depth for each segment. SANTS is not designed to reproduce this oracle at any cost. As Figure~\ref{fig:denoising-depth-diagnostic} shows, action error often saturates after moderate denoising. Even when additional video updates yield a small improvement, their marginal action benefit may not justify the added inference cost. The preferred terminal depth should therefore balance action quality against computation and may occur before the minimum-error endpoint. Although \(100\%\) remains the most frequent individual optimum, accounting for \(44.35\%\) of coarse and \(58.67\%\) of fine segments, intermediate depths collectively cover a substantial fraction of both groups. Fixed full denoising can consequently waste computation on low-value refinement and, for non-monotonic segments, may even produce a less useful action condition.

\begin{figure}[!t]
\centering
\includegraphics[width=0.88\linewidth]{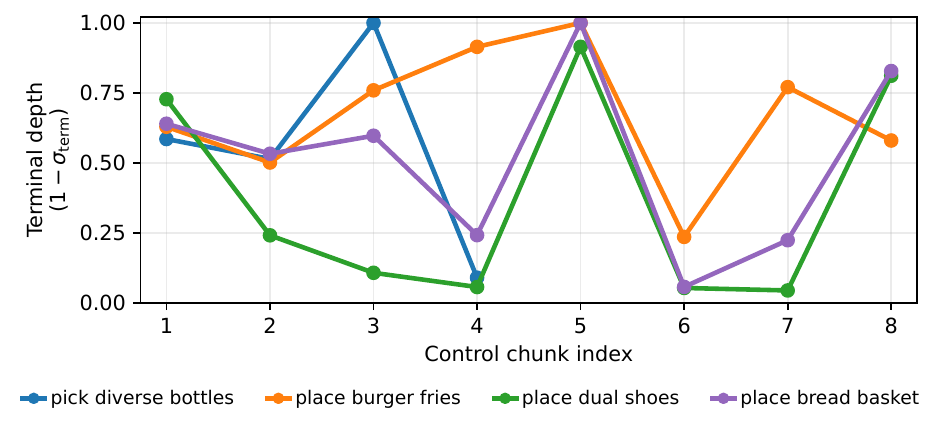}
\caption{Terminal denoising depths selected by SANTS across successive control segments in four closed-loop RoboTwin 2.0 rollouts. Each coloured line denotes one task rollout, and each marker denotes an action-chunk decision. Shorter curves reflect shorter original episode horizons.}
\label{fig:budget-trajectory}
\end{figure}

Finally, Figure~\ref{fig:budget-trajectory} visualizes the terminal denoising depths selected by SANTS during four closed-loop RoboTwin 2.0 rollouts. The selected depth varies across tasks and among successive control segments within the same rollout. Some segments stop after shallow denoising, whereas others continue close to full denoising. This result shows that the learned scheduler does not collapse to a fixed inference budget, but adjusts the amount of video computation according to the current state. The rollout results complement the offline diagnostic: the offline scan shows that action-useful video conditions can occur at different depths, while the closed-loop trajectories show that SANTS uses different depths during execution. The figure alone does not establish that every selected depth is optimal, because errors at the unselected depths are not evaluated. The ablations in Section~\ref{sec:exp-ablation} further test whether adaptive stopping and noise progression improve the closed-loop success--latency trade-off.

\begin{table}[!t]
\centering
\small
\setlength{\tabcolsep}{2.5pt}
\begin{tabular}{@{}lcccc@{}}
\toprule
& \multicolumn{3}{c}{RoboTwin 2.0 Success Rate \(\uparrow\)} & \\
\cmidrule(lr){2-4}
Method & Easy & Hard & Overall & \shortstack{A100 Latency\\(ms) \(\downarrow\)} \\
\midrule
LingBot-VA & 92.9\% & 91.5\% & 92.2\% & 2868.4 \\
Motus & 88.7\% & 87.0\% & 87.8\% & 1175 \\
Fast-WAM & 91.9\% & 91.8\% & 91.8\% & 190 \\
\(\pi_{0.5}\) & 82.7\% & 76.8\% & 79.8\% & 63 \\
SANTS & \textbf{94.6\%} & \textbf{94.2\%} & \textbf{94.4\%} & 523.7 \\
\bottomrule
\end{tabular}
\caption{Main comparison on RoboTwin 2.0. Inference latency is measured on A100 GPUs under the same evaluation pipeline.}
\label{tab:success-main}
\end{table}

\subsection{Experimental Setup}
\label{sec:exp-setup}

All experiments use a Wan2.2-5B-based video--action diffusion policy~\citep{wan2025wanopenadvancedlargescale} as the backbone for SANTS. The policy follows the interface in Section~\ref{sec:method-policy-interface}: the action branch conditions on the terminal intermediate video representation, and SANTS controls only the video denoising trajectory. During scheduler training, the video backbone, video head, and action head remain frozen, so differences between schedules can be attributed to inference-time noise control rather than changes in policy capacity.

Scheduler training uses multi-source simulated and real video--action trajectories, including public and self-collected real-robot data, and covers diverse manipulation scenes, task phases, and execution states. In simulation, we compare SANTS with LingBot-VA~\citep{li2026causalworldmodelingrobot}, Motus~\citep{bi2025motusunifiedlatentaction}, Fast-WAM~\citep{yuan2026fastwamworldactionmodels}, and \(\pi_{0.5}\)~\citep{intelligence2025pi05visionlanguageactionmodelopenworld}. For real-robot evaluation, we use \(\pi_{0.5}\) as the external VLA baseline and include a full-denoising WAM that shares the same video--action backbone with SANTS as a controlled pixel-space baseline. We report task success rate as the primary metric and end-to-end policy inference latency as the efficiency metric.

\subsection{Simulation and Real-Robot Results}
\label{sec:exp-success-rate}

We first evaluate task success and inference efficiency on RoboTwin 2.0, which covers bimanual fine manipulation, object contact, and randomized multi-task scenes. To isolate the effect of scheduling from downstream training budget, all methods use comparable fine-tuning budgets. SANTS starts from a pretrained video--action backbone, fine-tunes the backbone on RoboTwin 2.0 for \(30000\) optimization updates, and then freezes the video--action policy while fine-tuning only the adaptive scheduler for \(3000\) optimization updates. Baselines are fine-tuned from their pretrained checkpoints under the same total budget.

Table~\ref{tab:success-main} reports \(50\) trials on each of \(50\) RoboTwin 2.0 tasks. Across three SANTS seeds, overall success is \(94.1\%\), \(94.5\%\), and \(94.6\%\) (mean \(94.4\%\), std. \(0.3\%\)). SANTS reaches the highest overall success among the evaluated methods at \(523.7\) ms end-to-end latency. Compared with LingBot-VA, which uses the same Wan2.2 video weights with full denoising, SANTS improves success from \(92.2\%\) to \(94.4\%\) while reducing latency from \(2868.4\) to \(523.7\) ms. State-adaptive scheduling thus retains the pixel-space future-representation interface while removing much of the redundant video denoising cost.

\begin{table}[t]
\centering
\small
\renewcommand{\arraystretch}{0.94}
\setlength{\tabcolsep}{5pt}
\begin{tabular}{@{}lccc@{}}
\toprule
Task & \(\pi_{0.5}\) & Full WAM & SANTS \\
\midrule
\multicolumn{4}{@{}l}{\textit{AgileX}} \\
Cloth fold & 38.0\% & 54.0\% & \textbf{62.0\%} \\
Bag pack & 46.0\% & 68.0\% & \textbf{74.0\%} \\
Sock place & 60.0\% & 76.0\% & \textbf{78.0\%} \\
Charger ins. & 30.0\% & 38.0\% & \textbf{58.0\%} \\
\addlinespace[2pt]
\multicolumn{4}{@{}l}{\textit{UR10}} \\
Plate transf. & 68.0\% & 66.0\% & \textbf{80.0\%} \\
Fridge place & 48.0\% & 72.0\% & \textbf{74.0\%} \\
Fruit sort & \textbf{92.0\%} & 86.0\% & 86.0\% \\
\midrule
Mean SR \(\uparrow\) & 54.6\% & 65.7\% & \textbf{73.1\%} \\
Latency (ms) \(\downarrow\) & \textbf{63} & 2769.3 & 581.3 \\
\bottomrule
\end{tabular}
\caption{Real-robot results over \(50\) trials per task. Full WAM denotes full denoising; Mean SR is the unweighted mean over seven tasks, and latency is averaged across all tasks.}
\label{tab:real-robot-success}
\end{table}

We further evaluate real-robot performance on an AgileX dual-arm platform and a UR10 kitchen platform. The tasks cover bimanual coordination, deformable-object handling, precise insertion, container constraints, door interaction, and multi-object sorting. All methods use the same \(100\) hours of real-robot data and the same adaptation budget. We compare against \(\pi_{0.5}\) as an external VLA baseline and a Full-Denoising WAM that shares the same video--action backbone with SANTS but always denoises the future video representation to the final state. Each task is evaluated with \(50\) independent trials; Supplementary Section A provides the full hardware, reset, and evaluation protocol.

Across \(50\) trials per task, SANTS achieves \(73.1\%\) mean real-robot success, exceeding \(\pi_{0.5}\) and the Full-Denoising WAM by \(18.6\) and \(7.4\) percentage points, respectively. It matches or exceeds full denoising on every task while reducing mean end-to-end latency from \(2769.3\) to \(581.3\) ms, a \(79.0\%\) reduction. These results show that state-adaptive intermediate video representations provide a stronger success--latency tradeoff than full video denoising in real-robot execution.

\subsection{Ablation Study}
\label{sec:exp-ablation}

To isolate the two scheduling decisions in SANTS, we compare four controlled variants while keeping the video--action backbone, scheduler training budget, and evaluation protocol fixed. Fixed-5-step uses a latency-matched global five-step schedule, Fixed-full uses a fixed-step full-denoising schedule that always reaches \(\sigma_{\mathrm{term}}=0\), and Stop-only or Jump-only removes one adaptive component from SANTS. Reward-function ablations are reported in Supplementary Section D.

\begin{center}
\small
\renewcommand{\arraystretch}{0.92}
\setlength{\tabcolsep}{3.5pt}
\begin{tabular}{@{}lcccc@{}}
\toprule
Method & Stop & Jump & SR \(\uparrow\) & Lat. \(\downarrow\) \\
\midrule
Fixed-5-step & No & No & 87.2\% & 553.9 \\
Fixed-full & No & No & 89.8\% & 2769.3 \\
Jump-only & No & Yes & 91.6\% & 1137.7 \\
Stop-only & Yes & No & 92.7\% & 1329.3 \\
Full SANTS & Yes & Yes & \textbf{94.4\%} & \textbf{523.7} \\
\bottomrule
\end{tabular}
\captionof{table}{Scheduling ablations on RoboTwin 2.0. SR denotes overall success rate and Lat. denotes latency in milliseconds. Fixed-5-step uses a latency-matched global schedule.}
\label{tab:ablation-stop-jump}
\end{center}

Table~\ref{tab:ablation-stop-jump} shows that a fixed denoising budget cannot replace adaptive scheduling. The latency-matched Fixed-5-step baseline reaches only \(87.2\%\) success at \(553.9\) ms, whereas SANTS achieves \(94.4\%\) at \(523.7\) ms. This comparison is critical: SANTS improves success by \(7.2\) percentage points while using slightly lower latency, showing that its gain is not explained by simply reducing the number of video updates. Fixed-full also underperforms while being much slower, so always reaching \(\sigma_{\mathrm{term}}=0\) is both costly and suboptimal. Stop-only and Jump-only improve over fixed schedules but remain below Full SANTS: stopping alone cannot skip low-value regions after continuation, while jumping alone cannot choose the terminal representation used by the action branch. Terminal-state selection and relative progression are therefore complementary, consistent with Figure~\ref{fig:budget-trajectory}.

\section{Limitations}

Our evaluation is limited to RoboTwin 2.0 and two real-robot platforms, so broader tests across more embodiments, camera setups, task families, and video--action backbones are needed to establish generality. SANTS also keeps the video--action backbone frozen, leaving joint backbone--scheduler optimization unexplored. In deployment, it still depends on stopping and cost hyperparameters and mainly reduces video-denoising cost; extending adaptive budget allocation to action denoising, perception, communication, and control frequency remains future work.

\section{Conclusion}

This paper studies how pixel-space WAMs should choose future-video conditions at inference time. SANTS turns fixed video denoising into state-adaptive noise-trajectory control, stopping when the current intermediate representation is sufficient for action generation and otherwise predicting how far to progress along the remaining noise axis. Experiments show that SANTS reaches \(94.4\%\) success on RoboTwin 2.0 and \(73.1\%\) mean success on seven real-robot tasks, with \(523.7\) ms simulation latency and \(581.3\) ms all-task average real-robot latency, reducing full-denoising WAM latency by \(81.7\%\) and \(79.0\%\), respectively. Together with the stopping and progression ablations, these results suggest that WAM-style future reasoning need not rely on full pixel-level reconstruction; selecting the intermediate video state can preserve control benefits while removing redundant video inference.


%

\bibliography{references}
\clearpage

\appendix
\setcounter{figure}{0}
\setcounter{table}{0}
\setcounter{equation}{0}
\renewcommand{\thefigure}{S\arabic{figure}}
\renewcommand{\thetable}{S\arabic{table}}
\renewcommand{\theequation}{S\arabic{equation}}

\twocolumn[
  \begin{center}
    \vspace*{0.35in}
    {\LARGE\bf Supplementary Material for\\
    SANTS: A State-Adaptive Scheduler for World Action Models}
  \end{center}
  \vspace{0.75em}
]

%
%
%
%


\section{Real-Robot Evaluation Protocol}
\label{app:real-robot-protocol}

\begin{figure*}[t]
\centering
\includegraphics[width=0.95\textwidth]{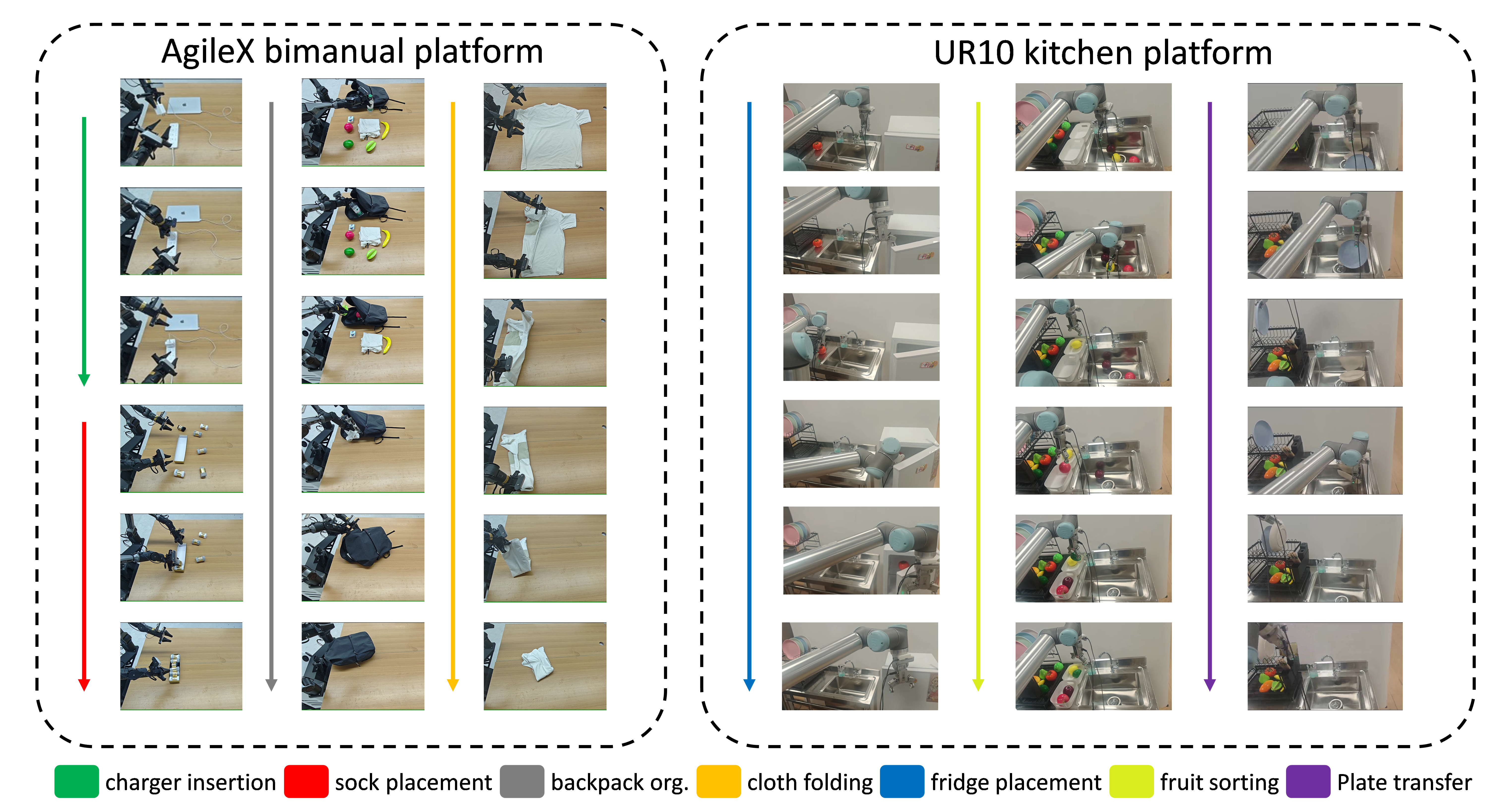}
\caption{Real-robot task sequences on the AgileX bimanual and UR10 kitchen platforms. Colored arrows indicate temporal progress for each task.}
\label{fig:app-real-robot-task-sequences}
\end{figure*}

This appendix provides the hardware, observation, control, randomization, and evaluation protocol for the real-robot experiments. The evaluation covers seven tasks on two platforms: four tasks on a Songling/AgileX bimanual platform and three tasks on a UR10 kitchen platform. For each task and each method, we run \(50\) independent trials. Each trial starts from a valid manually reset initial state and terminates when the task succeeds, reaches the time limit, triggers a safety stop, or enters an unrecoverable failure state. The three methods, \(\pi_{0.5}\), Full-Denoising WAM, and SANTS, use the same task definitions, workspace bounds, camera inputs, action interface, deployment stack, and termination rules. They also share the same \(100\) hours of real-robot data and the same adaptation budget. Methods are evaluated under the same or equivalent initial-state distributions, and no method uses additional demonstrations, privileged state information, or a different reset distribution.

The Songling/AgileX experiments use the Cobot Magic bimanual kit and its associated vision system. The platform provides two wrist-camera views and one top-down third-person view, with Orbbec Dabai depth cameras used to observe the two end effectors, task objects, and the global task region. The four tasks on this platform evaluate cloth folding, backpack organization, sock placement, and charger insertion, covering bimanual coordination, deformable-object manipulation, local object arrangement, and precise insertion. The policy outputs bimanual end-effector motion and gripper commands. Demonstrations are exported as synchronized video-action sequences at \(15\) Hz; real-robot evaluation uses the same action time base and queries the model at action-chunk boundaries. During evaluation, we randomize task-object positions, orientations, and object colors for replaceable items, keep background objects fixed, and repeat trials under different lighting conditions.

The UR10 kitchen platform consists of a UR10 arm, an OnRobot RG2-FT gripper, a third-person camera, and a wrist camera. Both cameras are Intel RealSense D435i devices. The three tasks on this platform evaluate plate transfer, refrigerator placement, and fruit sorting, covering object transfer, articulated-door interaction, container constraints, and multi-object classification. The UR10 policy interface uses end-effector pose and gripper state as actions. Demonstrations record robot state, gripper state, and two-view videos under the same action parameterization. The PICO Ultra 4 VR teleoperation client reads controller commands at \(60\) Hz and produces robot targets; during recording, two-view videos, robot state, and gripper state are synchronized and saved at a \(30\) Hz dataset clock. The UR10 trajectories used for adaptation are resampled to \(15\) Hz. At real-robot inference time, the client sends predicted actions at \(10\) Hz. Each model query returns \(16\) future action steps, corresponding to an execution window of approximately \(1.6\) s, after which the policy replans. Across trials, we randomize the positions, orientations, and colors of target objects such as fruits, while keeping background objects and major furniture fixed. Evaluation is repeated under different lighting conditions to test closed-loop robustness under real visual variation.

Before each trial, the robot returns to a fixed home pose and the scene is manually reset according to the task-specific randomization protocol. Human intervention is limited to pre-trial scene reset and post-trial safety recovery; no online correction or mid-trial teleoperation is allowed. Success is determined from the final task state: the target object must reach the specified region or pose, and the execution must not trigger a safety stop, object drop, incorrect placement, or irrecoverable collision. All trials that do not satisfy the success condition are counted as failures under the same task-level termination rules. Failure-type labels are not used for training, model selection, or result aggregation.

\section{Scheduler Implementation and Hyperparameters}
\label{app:scheduler-implementation}

This appendix reports the implementation details needed to reproduce SANTS. The main paper defines the cumulative stopping hazard and relative noise progression; here we summarize the plug-in scheduler, the target-domain tuning setting, and the fixed hyperparameters used in the main experiments.

\paragraph{Plug-in scheduler.}
SANTS uses a lightweight scheduler that is independent of the WAM backbone. At each video decision point, it reads the current video-state representation \(z_k\) and noise level \(\sigma_k\), and predicts a stopping hazard together with a relative noise-progression ratio. The scheduler does not update the video backbone, video head, action head, or action denoising process. A trained SANTS scheduler can therefore be attached to compatible WAM inference pipelines that expose intermediate video features and support nonuniform video noise updates.

Because the scheduler is small, it learns a compact decision rule tied to the task-phase distribution of the source domain. Direct transfer is already effective, but a short target-domain tuning stage provides a more stable match to the visual states and action demands of the target platform. In our real-robot adaptation setting, we tune only the scheduler for \(3000\) optimization steps and keep all WAM parameters frozen. Table~\ref{tab:app-scheduler-transfer} shows that direct transfer reaches \(70.6\%\) average success over seven real-robot tasks, while target-domain tuning improves it to \(73.1\%\).

\begin{table}[t]
\centering
\small
\setlength{\tabcolsep}{6pt}
\begin{tabular}{lcc}
\toprule
Setting & Tuning steps & Avg. SR \(\uparrow\) \\
\midrule
Direct transfer & \(0\) & 70.6\% \\
Target-tuned & \(3000\) & \textbf{73.1\%} \\
\bottomrule
\end{tabular}
\caption{Scheduler transfer on real robots.}
\label{tab:app-scheduler-transfer}
\end{table}

\paragraph{Architecture and hyperparameters.}
The scheduler is attached after the frozen WAM video branch. It contains a two-layer MLP trunk and two output heads. The input concatenates the pooled video feature \(z_k\) with the current noise level \(\sigma_k\). The two heads predict the stopping signal and the relative noise-progression policy, respectively. Table~\ref{tab:app-scheduler-impl} lists the SANTS network size, noise thresholds, and stopping threshold. These hyperparameters are selected on a held-out validation split and then fixed across tasks.

\begin{table}[t]
\centering
\small
\setlength{\tabcolsep}{5pt}
\begin{tabular}{ll}
\toprule
Item & Value \\
\midrule
Input dim. & \(D_{\mathrm{wam}}+1\) \\
MLP layers & \(2\) \\
hidden width & \(512\) \\
bottleneck & \(256\) \\
Output heads & \(2\) \\
Start noise \(\sigma_{\mathrm{start}}\) & \(1.00\) \\
Early-stop noise \(\sigma_{\mathrm{early}}\) & \(0.963\) \\
Stop threshold \(\eta\) & \(0.85\) \\
Forced terminal \(\sigma_{\min}\) & \(0.01\) \\
Full-denoise ref. \(\sigma_{\mathrm{full}}\) & \(0\) \\
\bottomrule
\end{tabular}
\caption{SANTS scheduler architecture and hyperparameters.}
\label{tab:app-scheduler-impl}
\end{table}

\section{Reinforcement-Learning Post-Training and Reward Definition}
\label{app:reward-details}

\paragraph{Overview.}
This section expands the compact path-return form in Eq.~\eqref{eq:sants-reward-compact} of the main paper. We first define the action-quality gain and the video-budget cost, and then describe the PPO post-training configuration used for the scheduler.

\paragraph{Action-Quality Error.}
The reward does not use video sharpness. Instead, it compares the action \(\hat a_{\tau}\) generated from a scheduled path \(\tau\) with the demonstration action \(a^\star\) in the physical action space. We use two complementary errors. The sequence action error \(L_{\mathrm{seq}}\) measures deviations in future end-effector position, orientation, and gripper state, while the temporal-difference action error \(L_{\Delta}\) measures whether the motion trend between adjacent actions matches the demonstration trajectory. Position error is computed in the physical displacement units recorded by the dataset, orientation error uses quaternion angular distance, and gripper state is mapped to \([0,1]\) before applying binary cross entropy. In implementation, we upweight timesteps near large end-effector motions and gripper events. A gripper event is detected when either the left or right gripper changes by more than \(0.25\).

\paragraph{Two-Anchor Quality Normalization.}
Using the absolute action error directly would entangle sample difficulty with scheduling quality. For each input, we therefore run two additional reference paths: a shallow-denoising path \(\tau_{\mathrm{lo}}\) and a full-denoising path \(\tau_{\mathrm{hi}}\), with terminal noise levels \(\sigma_{\mathrm{lo}}\approx0.963\) and \(\sigma_{\mathrm{hi}}=0\), respectively. These paths produce actions \(\hat a_{\mathrm{lo}}\) and \(\hat a_{\mathrm{hi}}\). For either error type \(m\in\{\mathrm{seq},\Delta\}\), we define \(G_m\) as the recoverable error gap between shallow and full denoising, with a floor \(\epsilon\) to avoid degenerate denominators:
\[
G_m=\max\!\left(L_m(\hat a_{\mathrm{lo}},a^\star)-L_m(\hat a_{\mathrm{hi}},a^\star),\epsilon\right).
\]
The normalized gain of the current scheduled path relative to the shallow-denoising reference is
\begin{equation}
u_m(\tau)=
\mathrm{clip}_{[-1,1]}\!
\left(
\frac{
L_m(\hat a_{\mathrm{lo}},a^\star)-
L_m(\hat a_{\tau},a^\star)}
{G_m}
\right).
\label{eq:app-anchor-gain}
\end{equation}
When shallow and full denoising differ only marginally, further video inference provides little recoverable action information. To prevent such samples from dominating training, we introduce a difficulty coefficient
\begin{equation}
D_m=
\frac{
\max\!\left(L_m(\hat a_{\mathrm{lo}},a^\star)-
L_m(\hat a_{\mathrm{hi}},a^\star),0\right)}
{
\max\!\left(L_m(\hat a_{\mathrm{lo}},a^\star)-
L_m(\hat a_{\mathrm{hi}},a^\star),0\right)+\kappa}.
\label{eq:app-difficulty-gate}
\end{equation}
The final quality term rewards positive recovery while penalizing scheduled paths that perform worse than the shallow-denoising reference:
\begin{equation}
\begin{aligned}
Q_m(\tau)={}&w_m\Bigl[
D_m\max(u_m(\tau),0)\\
&\quad-\lambda_{\mathrm{pen}}\max(-u_m(\tau),0)
\Bigr],\\
Q(\tau)={}&Q_{\mathrm{seq}}(\tau)+Q_{\Delta}(\tau).
\end{aligned}
\label{eq:app-quality-gain}
\end{equation}
This design focuses the reward on whether the current path recovers the action-relevant information made available by full denoising while terminating at a higher noise level. It therefore avoids simply encouraging all states to progress to \(\sigma_{\mathrm{term}}=0\).

\paragraph{Quality-Cost Path Return.}
Let \(n_\tau\) denote the number of video-state updates actually executed by path \(\tau\), and let \(N_{\mathrm{full}}\) denote the number of updates in the full-denoising reference trajectory. We use the following normalized monotonic cost:
\begin{equation}
C(n_\tau)=
\lambda_c
\frac{\sum_{i=1}^{n_\tau} c_i}{\sum_{i=1}^{N_{\mathrm{full}}} c_i},
\qquad
c_i=c_0+c_1\left(\frac{i}{N_{\mathrm{full}}}\right)^{\gamma}.
\label{eq:app-update-cost}
\end{equation}
The path return is then
\begin{equation}
R(\tau)=Q(\tau)-C(n_\tau).
\label{eq:app-expanded-reward}
\end{equation}
This is equivalent to Eq.~(8) in the main paper, where \(Q(\tau)\) is the weighted action-quality gain and \(C(n_\tau)\) is the video-update cost.

\paragraph{Path-Level PPO Post-Training.}
During post-training, we update only the SANTS scheduler head and freeze the WAM backbone, the video head, the action head, and the action-denoising process. Each optimization step samples one rollout from the current scheduler policy and additionally runs two no-gradient anchor rollouts with fixed terminal noise levels, \(\sigma_{\mathrm{lo}}\approx0.963\) and \(\sigma_{\mathrm{hi}}=0\). These anchors are used only for within-input reward normalization and are not treated as additional training targets.

For PPO updates, we recompute the current policy probability on the sampled path and form the path log-probability from both the stop/continue decisions and the jump density when continuation is selected. The advantage is computed as the path return minus an EMA baseline with decay \(0.95\), without cross-sample reward normalization. We use AdamW~\citep{loshchilov2017decoupled} with learning rate \(10^{-4}\) and weight decay \(0.01\). Each sampled path is optimized for \(4\) PPO epochs, with clip range \(0.2\) and global-norm gradient clipping at \(1.0\). Reference regularization includes a KL term between the stopping distribution and the full-denoising reference trajectory, as well as a KL term between the jump Beta distribution and the reference progression ratio. The corresponding weight is \(\lambda_{\mathrm{KL}}=0.02\), and the jump log-probability scale is \(0.5\). Training uses no entropy bonus and no early stopping based on training reward. The deployed checkpoint is selected on the validation split using task success rate, path return, and the average number of video updates.

\paragraph{Training and Deployment Summary.}
During training, we freeze the video--action backbone and update only the scheduler head. For each demonstration segment, we sample one trajectory from the current policy, run two fixed-terminal-noise anchors, and apply path-level PPO post-training with the resulting return. During validation, we evaluate success rate, path return, and the average number of video updates on a held-out split, and select the deployed checkpoint accordingly. During deployment, the scheduler reads the current video representation \(z_k\) and noise level \(\sigma_k\) at each video decision point, then decides whether to continue video updates or terminate video generation according to the stopping threshold and noise-progression ratio.

\paragraph{Hyperparameter Selection and Sensitivity.}
All reward and PPO hyperparameters are selected on the held-out validation split and then fixed globally in the main experiments. We do not tune hyperparameters separately for individual RoboTwin tasks or real-robot tasks. Table~\ref{tab:app-rl-hyperparams} summarizes the global configuration used for path rewards and PPO post-training. Empirically, success rate and latency are most sensitive to the stopping threshold \(\eta\), the early-stop reference noise \(\sigma_{\mathrm{early}}\), the compute-cost weight \(\lambda_c\), and the difficulty smoothing term. Action-error weights and key-frame weights mainly affect training stability. Around the selected values, the learned policy remains qualitatively consistent, but an overly large compute-cost weight can make the scheduler terminate prematurely during contact and alignment phases.

\begin{table*}[t]
\centering
\small
\setlength{\tabcolsep}{8pt}
\begin{tabular*}{0.94\textwidth}{@{\extracolsep{\fill}}lll@{}}
\toprule
Category & Hyperparameter & Value \\
\midrule
Reward anchors & \(\sigma_{\mathrm{lo}},\sigma_{\mathrm{hi}},\sigma_{\mathrm{ref}}\) & \(0.963,0,0\) \\
Quality gain & \(w_{\mathrm{seq}},w_{\Delta}\) & \(0.6,0.4\) \\
Sequence-error weights & pos, rot, gripper & \(1.0,0.6,1.0\) \\
Difference-error weights & pos, rot, gripper & \(1.0,0.2,1.0\) \\
Key-frame weights & \(\rho,\eta_{\mathrm{mot}},\eta_{\mathrm{grip}},\eta_{\mathrm{early}}\) & \(0.06,0.8,1.8,0\) \\
Gripper-event threshold & \(|\Delta g|\) & \(0.25\) \\
Penalty and cost & \(\lambda_{\mathrm{pen}},\lambda_c\) & \(1.0,0.25\) \\
Cost schedule & \(c_0,c_1,\gamma\) & \(0.2,0,2\) \\
Normalization stabilizers & gap floor, difficulty \(\kappa\) & \(10^{-3},0.02\) \\
PPO optimizer & lr, weight decay & \(10^{-4},0.01\) \\
PPO update & epochs, clip, grad norm & \(4,0.2,1.0\) \\
Policy regularization & \(\lambda_{\mathrm{KL}}\), jump log-prob scale & \(0.02,0.5\) \\
Advantage estimation & EMA decay, reward normalization & \(0.95,\) none \\
\bottomrule
\end{tabular*}
\caption{Reward and PPO post-training hyperparameters. All values are selected on the held-out validation split and then fixed for all experiments.}
\label{tab:app-rl-hyperparams}
\end{table*}

\section{Reward-Function Ablations}
\label{app:additional-ablations}
\label{app:ablation-reward}

Table~\ref{tab:ablation-stop-jump} in the main paper isolates the contributions of the two scheduling decisions, stopping and noise progression. Here, we do not introduce additional stopping variants or hyperparameter scans. Instead, we test whether the main components of the path-level reward are necessary. All variants use the same frozen video--action backbone, scheduler architecture, training split, post-training budget, and evaluation protocol. Except for the reward term that is removed or replaced, all settings are kept fixed. We report both success rate and end-to-end latency to separate action-quality gains from changes in compute allocation.

\begin{table*}[t]
\centering
\small
\setlength{\tabcolsep}{8pt}
\begin{tabular*}{0.94\textwidth}{@{\extracolsep{\fill}}llcc@{}}
\toprule
Variant & Removed or replaced component & Overall SR \(\uparrow\) & Latency (ms) \(\downarrow\) \\
\midrule
Full path reward & none & 94.4\% & 523.7 \\
Raw action MSE reward & anchor-normalized gain & 89.7\% & 436.3 \\
No temporal-difference error & \(L_{\Delta}\) term & 93.6\% & 576.4 \\
No difficulty gate & \(D_m\) gate & 90.6\% & 489.6 \\
No update-cost penalty & \(C(n_{\tau})\) term & 94.9\% & 2197.8 \\
\bottomrule
\end{tabular*}
\caption{Path-reward ablations. Except for the reward definition, all variants use the same frozen video--action backbone, scheduler architecture, training budget, and evaluation protocol.}
\label{tab:app-ablation-reward}
\end{table*}

Table~\ref{tab:app-ablation-reward} shows that different components of the path-level reward affect control quality and compute allocation in distinct ways. Replacing the two-anchor normalized gain with raw action MSE reduces success from \(94.4\%\) to \(89.7\%\), although latency decreases to \(436.3\) ms. This indicates that directly optimizing unnormalized action error can produce a cheaper termination policy whose control quality is insufficient. Removing the temporal-difference error slightly lowers success to \(93.6\%\) and increases latency to \(576.4\) ms, suggesting that \(L_{\Delta}\) mainly helps the reward capture motion trends and contact transitions rather than merely reducing computation. Removing the difficulty gate decreases success to \(90.6\%\), which indicates that positive gains must be suppressed on samples where the full-denoising anchor itself provides little improvement; otherwise, the reward can be miscalibrated by low-information states. Finally, removing the update-cost penalty gives a similar, and slightly higher, success rate of \(94.9\%\), but increases latency to \(2197.8\) ms, which is \(4.2\times\) that of the full reward. The full path reward is therefore not designed to maximize a single success number in isolation. It reduces end-to-end latency by \(76.2\%\) relative to the no-cost variant while losing only \(0.5\) percentage points in success, supporting the need to jointly model action-quality gains and computational cost.

%


\end{document}